\documentclass[10pt,twocolumn,letterpaper]{article}

\usepackage{wacv}
\usepackage{times}
\usepackage{epsfig}
\usepackage{graphicx}
\usepackage{amsmath}
\usepackage{amssymb}
\usepackage{booktabs}
\usepackage{amsmath,amssymb,amsfonts}
\usepackage{makecell}
\usepackage{hhline}
\usepackage{multirow}
\usepackage{authblk}
\usepackage{marvosym}
\usepackage{xcolor}

\def\wacvPaperID{****} 

\wacvfinalcopy 

\ifwacvfinal
\usepackage[breaklinks=true,bookmarks=false]{hyperref}
\else
\usepackage[pagebackref=true,breaklinks=true,colorlinks,bookmarks=false]{hyperref}
\fi

\begin{document}

\title{DuAT: Dual-Aggregation Transformer Network for Medical Image Segmentation}  

\author[1]{Feilong Tang}
\author[1]{Qiming Huang}
\author[1]{Jinfeng Wang}
\author[2]{Xianxu Hou}
\author[1]{Jionglong Su\textsuperscript{\Letter}}
\author[1]{Jingxin Liu\textsuperscript{\Letter}}
\affil[1]{School of AI and Advanced Computing, Xi’an Jiaotong-Liverpool University, Suzhou, China}
\affil[2]{School of Computer Science and Software Engineering, Shenzhen University, Shenzhen, China}

\maketitle
\thispagestyle{empty}

\begin{abstract}
    Transformer-based models have been widely demonstrated to be successful in computer vision tasks by modelling long-range dependencies and capturing global representations. 
    However, they are often dominated by features of large patterns leading to the loss of local details (e.g., boundaries and small objects), 
    which are critical in medical image segmentation. To alleviate this problem, we propose a Dual-Aggregation Transformer Network called DuAT, which is characterized by two innovative designs, 
    namely, the Global-to-Local Spatial Aggregation (GLSA) and Selective Boundary Aggregation (SBA) modules. 
    The GLSA has the ability to aggregate and represent both global and local spatial features, which are beneficial for 
    locating large and small objects, respectively. The SBA module is used to aggregate the boundary characteristic from 
    low-level features and semantic information from high-level features for better preserving boundary details and locating
    the re-calibration objects. Extensive experiments in six benchmark datasets demonstrate that our proposed model 
    outperforms state-of-the-art methods in the segmentation of skin lesion images, and polyps in colonoscopy images. 
    In addition, our approach is more robust than existing methods in various challenging situations such as small object 
    segmentation and ambiguous object boundaries.
\end{abstract}

\section{Introduction}
\label{sec:intro}

Medical image segmentation is a computer-aided automatic procedure for extracting the region of interest (RoI), e.g., tissues, lesions, and body organs. It can assist clinicians by making diagnostic and treatment processes more efficient and precise. For instance, colonoscopy is the gold standard for detecting colorectal lesions, and accurately locating early polyps is of great significance for clinical prevention of rectal cancer \cite{favoriti2016worldwide}. Likewise, melanoma skin cancer is one of the most rapidly increasing cancers worldwide. Segmentation of skin lesions from dermoscopic images is a critical step in skin cancer diagnosis and treatment planning \cite{mathur2020cancer}. However, it is impractical to manually annotate these structures in clinical practice due to the tedious, time-consuming, and error-prone process. There is a growing need for automatic as well as accurate image segmentation.
\begin{figure}[t]
   \centerline{\includegraphics[width=\linewidth]{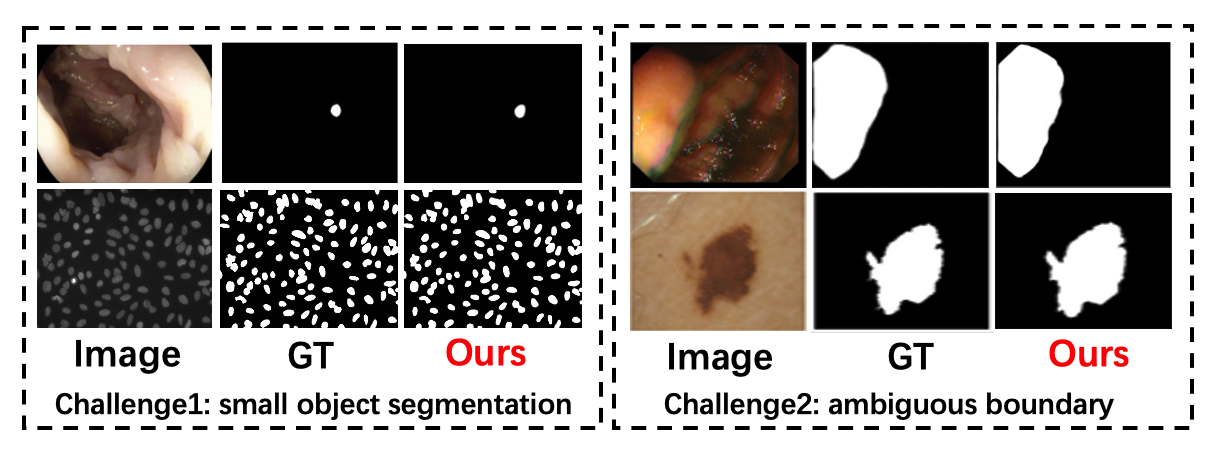}}
   \caption{
   Main challenges of medical images. \textbf{Challenge1}: Small object segmentation are difficult to segment due to their low contrast and strong camouflage. \textbf{Challenge2}: The object boundary for medical image is ambiguous due to image acquisition influence.}
   \label{fig555}
\end{figure}
With the rapid development of deep learning, an increasing number of deep convolutional neural networks (DCNNs) \cite{fan2020pranet,fu2018joint,Unet,zhang2020adaptive,zhou2018unet++} are proposed for medical image segmentation. The limitations of the receptive field in DCNNs make it difficult to capture the global representation. To solve this problem, dilated convolution \cite{chen2017deeplab, yu2015multi,zhang2020dynamic} is proposed for semantic segmentation task. Furthermore, attention models \cite{wang2018non, huang2019ccnet,zhang2019dual} are developed to better capture long-range context information. These alternatives achieve promising results in semantic segmentation. 

Transformer-based methods \cite{beal2020toward, carion2020end, chen2021pre,han2021transformer,touvron2021training,zheng2021rethinking, wang2022pvt} have been proposed and achieved comparable performance to state-of-the-art results. Transformer is originally used for sequence-to-sequence predictive modelling in natural language processing (NLP) tasks with pure attention structure that is good at capturing long-range dependencies \cite{vaswani2017attention}. Vision Transformer-based methods \cite{beal2020toward, carion2020end, chen2021pre,han2021transformer,touvron2021training,zheng2021rethinking, wang2022pvt} have also been proposed and demonstrate promising performance. ViT \cite{dosovitskiy2020image} constructs a vector sequence by dividing each image into fixed-size patches, subsequently applies a multi-head self-attention (MHSA) and the Multilayer Perceptron (MLP) structure which demonstrate the advanced learning ability for long-distance feature dependence. The recent works \cite{dong2021polyp, xie2021segformer, wang2022stepwise} demonstrate that the pyramid structure is applicable in Transformers and more suitable for various downstream tasks. Unfortunately, capturing long-range dependencies destroys part of local features, which could result in overly smooth predictions for small objects and blurred boundaries between objects.

Therefore, building a model that retains both local and global features remains challenging. Li \textit{et al.} \cite{li2019global} explore the local context for the aggregated long-range relationship to be more accurately distributed in local regions. Recently, some hybrid architectures of Transformer and CNN are proposed \cite{chen2021transunet,ranftl2021vision,zheng2021rethinking}, which aim to combine the advantages of both models. Wang \textit{et al}. \cite{wang2022stepwise} propose progressive locality decoder to emphasize local features and restrict attention dispersion. Chen \textit{et al}. \cite{chen2021transunet} propose TransUnet, which utilizes the underlying features of CNNs and subsequently uses the transformers to model global interactions to strengthen local features. However, feeding local information directly into the transformer cannot precisely handle local context relationships, resulting in the local information being overwhelmed by the dominant global context. Ultimately it leads to  inferior results in the medical image segmentation of small objects.

Considering the problem of unclear object boundary information in semantic segmentation. Previous studies \cite{li2020gated,chen2018encoder,dong2021polyp} explore fusing low-scale boundary information and high-scale semantic information  to better preserve boundary details. Moreover, Zhang \textit{et al.} \cite{zhang2018exfuse} incorporate semantic information into low-level features while embedding more spatial information into high-level features to make up for the semantic and resolution gap between feature maps. Takikawa \textit{et al.} \cite{takikawa2019gated} and Zhen \textit{et al.} \cite{zhen2020joint} design a boundary stream and couple the task of boundary and semantics modeling. Inspired by \cite{zhang2018exfuse}, we exploit the boundary information to selectively fuse it with the high-level semantic features, which will further enhance the semantic representations rather than simply combining them. 

In this paper, we propose a novel pyramid transformer for medical image segmentation, referred to as the Dual-Aggregation Transformer Network (DuAT), consisting of the Global-to-Local Spatial Aggregation (\textit{GLSA}) to combine local and global features, as well as the Selective Boundary Aggregation (\textit{SBA}) module to enhance the boundary information and locating the re-calibration objects. We believe that global spatial features help locate large objects, and local spatial features are crucial for identifying small ones. Finally, boundary information is aggregated to fine-tune object boundaries and re-calibrate coarse predictions, Specifically, the \textit{GLSA} module is responsible for extracting and fusing local and global spatial features from the backbone. We separate the channels, one for global representation extracted by Global context (GC) block \cite{cao2019gcnet}, and the other for local information extracted by multiple depth-wise convolutions. The \textit{SBA} module aims to simulate the biological visual perception process, distinguishing objects from background. Specifically, it incorporates shallow- and deep-level features to establish the relationship between body areas and boundary, enhancing the boundaries characteristics. Our experimental results demonstrate three advantages of our model: more lightweight, better learning ability and improved generalization capability than previous state-of-the-art approaches. 

In summary, the main contribution of this paper is three-fold.
\begin{itemize}
\item[$\bullet$] We propose a novel framework, named Dual-Aggregation Transformer Network (\textbf{DuAT}), for medical images segmentation, which adapts the pyramid vision transformer as encoder to extract more robust features than the existing CNN-based methods.
\end{itemize}

\begin{itemize}
\item[$\bullet$] We design dual aggregation modules,  Global-to-Local Spatial Aggregation (\textbf{GLSA}) module and  Selective Boundary Aggregation (\textbf{SBA}) module. Their purpose is to solve two challenges that are intuitively demonstrated in Figure \ref{fig555}. Specifically, the GLSA module simultaneously extract local spatial detail information and global spatial semantic information, which reduces incorrect information in the high-level features. The SBA module aims to fine-tune object boundaries, which can well address the ``ambiguous" problem of boundaries.
\end{itemize}

\begin{itemize}
\item[$\bullet$] Extensive experiments on five polyp datasets (ETIS \cite{vazquez2017benchmark}, CVC-ClinicDB \cite{silva2014toward}, CVC-ColonDB \cite{bernal2015wm}, EndoScene-CVC300 \cite{tajbakhsh2015automated}, Kvasir \cite{jha2020kvasir}), skin lesion dataset (ISIC-2018 \cite{codella2019skin}) and  2018 Data Science Bowl \cite{caicedo2019nucleus} demonstrate that the proposed DuAT methods advances the state-of-the-art (SOTA) performance.
\end{itemize}

\begin{figure*}[!t]
   \centerline{\includegraphics[width=\linewidth]{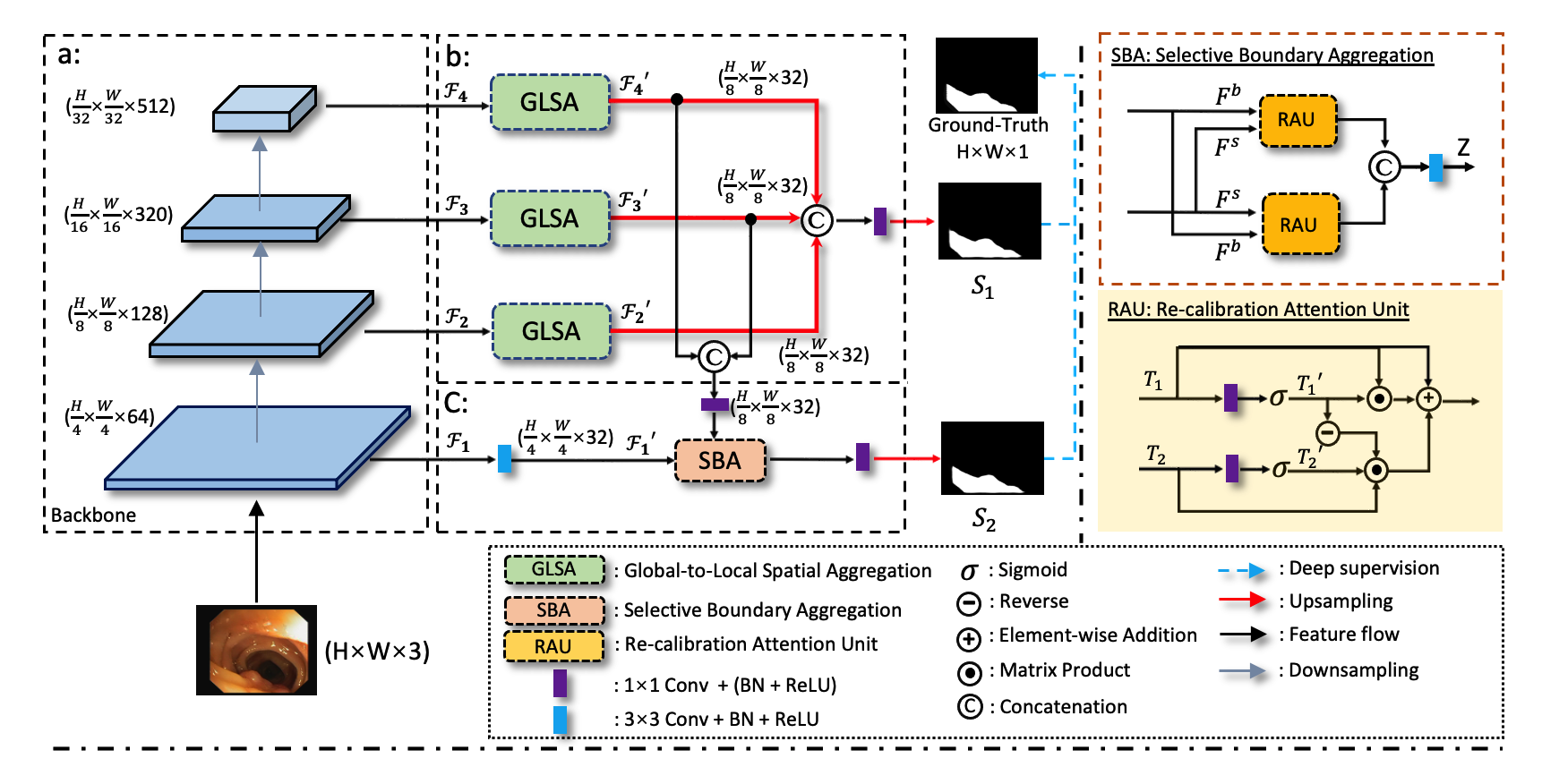}}
   \caption{The overall architecture of Dual-Aggregation Transformer Network (DuAT). The entire model is divided into three parts: (a) pyramid vision transformer (PVT) as backbone; (b) pyramid Global-to-Local Spatial Aggregation (GLSA) Module; (c) Selective Boundary Aggregation (SBA) module and it shown on the red box.} 
   \label{fig1}
\end{figure*}

\section{Related Work}

\subsection{Vision Transformer}

Transformer has dominated the field of NLP with its MHSA layer to capture the pure attention structure of long-range dependencies. Different from convolutional layer, MHSA layer has dynamic weight and global receptive field, which makes it more flexible and effective. Dosovitskiy \textit{et al.} propose a vision transformer (ViT) \cite{dosovitskiy2020image}, which is an end-to-end model using the Transformer structure for image recognition task. Specifically, it divides an image into fixed-size patches, which are sequentially fed to multiple Transformer encoder blocks to model the patches. In addition, previous work has proved that the pyramid structure in convolutional networks is also suitable for Transformer and various downstream tasks, such as Swin Transformer \cite{liu2021swin}, PVT \cite{wang2022pvt}, Segformer \cite{xie2021segformer}, etc. PVT requires less computation than ViT and adopts the classical Semantic-FPN to deploy the task of semantic segmentation. 

In medical image segmentation, TransUNet \cite{chen2021transunet} demonstrates that Transformer can be used as powerful encoders for medical image segmentation. TransFuse \cite{zhang2021transfuse} is proposed to improve efficiency for global context modeling by fusing transformers and CNNs. Furthermore, to train the model effectively on medical images, Polyp-PVT \cite{dong2021polyp} introduces Similarity Aggregation Module based on graph convolution domain \cite{lu2019graph}. Inspired by these approaches, we propose a new transformer-based medical segmentation framework, which can accurately locate small objects.

\begin{figure*}[!t]
   \centerline{\includegraphics[width=\linewidth]{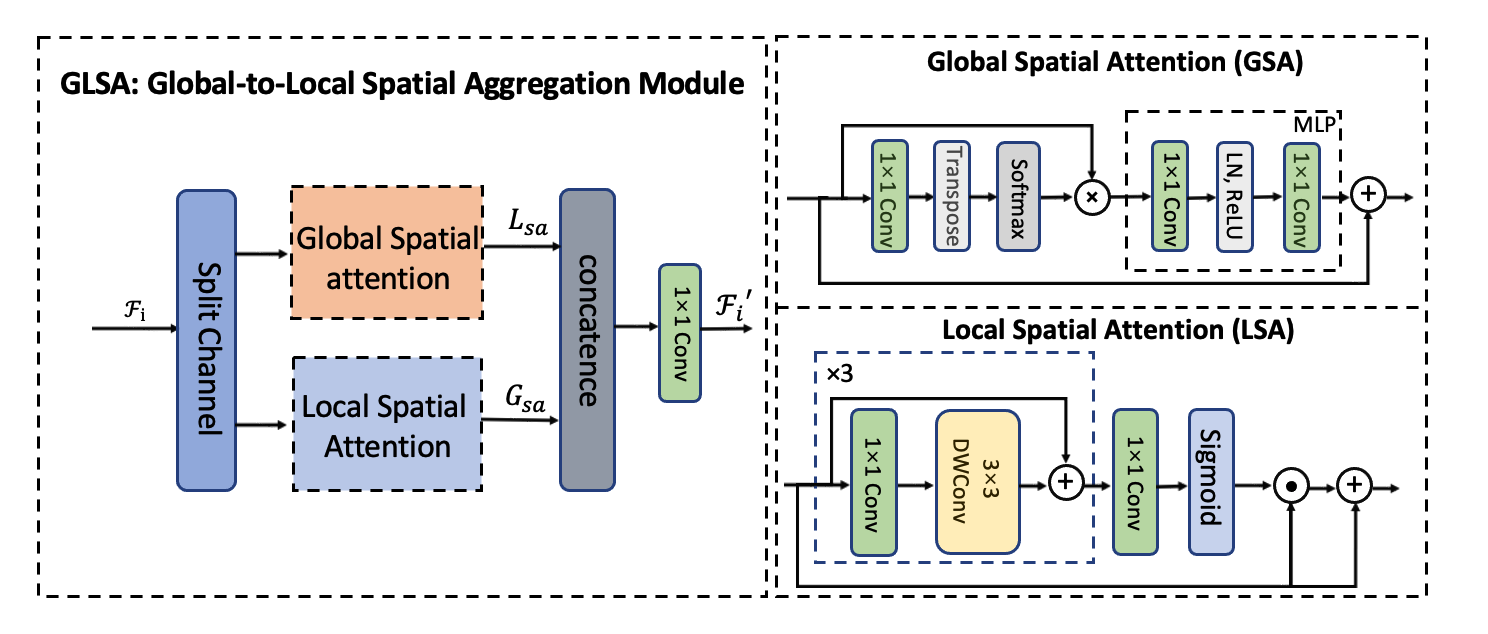}}  
   \caption{
       Overview of the Global-to-Local Spatial Aggregation Module \textit{GLSA}, it is composed of global spatial attention (\textit{GSA}) and local spatial attention \textit{(LSA)}.}
   \label{fig2}
\end{figure*}

\subsection{Image Boundary Segmentation} 

Recently, learning additional boundary information has shown superior performance in many image segmentation tasks. In the early research on FCN-based semantic segmentation, Bertasius \textit{et al.} \cite{bertasius2016semantic} and Chen \textit{et al.} \cite{chen2017deeplab} use boundaries for post-starting to refine the result at the end of the network. Recently, several approaches explicitly model boundary detection as an independent sub-task in parallel with semantic segmentation for sharper result. Ma \textit{et al.} \cite{ma2021boundary} explicitly exploit the boundary information for context aggregation to further enhance the semantic representation of the model. Li \textit{et al.} \cite{li2020improving} point out that the object boundary and body parts correspond to the high-frequency and low-frequency information of an image, respectively, based on which they decouple the body and edge with diverse supervisions. Ji \textit{et al.} \cite{ji2022fast} fuse the low-level edge-aware features and constraint it with the explicit edge supervision. Different from the above works, we propose a novel aggregation method to achieve more accurate localisation and boundary delineation of objects.

\section{Method}

As given in Figure \ref{fig1}, our DuAT model consists of the following:- a pyramid vision transformer (PVT) encoder, a \textit{SBA} module, and \textit{GLSA} module.

\subsection{Transformer Encoder} Some recent studies \cite{bhojanapalli2021understanding,xie2021segformer} report that vision transformers \cite{dosovitskiy2020image,wang2022pvt} have stronger performance and robustness to input disturbances (e.g., noise) than CNNs. Inspired by this, we use the Transformer based on pyramid structure as the encoder. Specifically, the pyramid vision transformer (PVT) \cite{wang2022pvt} is utilized as the encoder module for multi-level feature maps \{${\mathcal{F}_{i}|i \in (1,2,3,4) }$\} extraction. Among these feature maps, $\mathcal{F}_{1}$ gives detailed boundary information of target, and $\mathcal{F}_{2}$, $\mathcal{F}_{3} $ and $ \mathcal{F}_{4} $ provide high-level features.




\subsection{Selective Boundary Aggregation} 
As observed in \cite{zhang2018exfuse,li2020improving}, shallow- and deep-layer features complement each other. The shallow layer has less semantics but is rich in details, with more distinct boundaries and less distortion. Furthermore, the deep level contains a rich semantic information. Therefore, directly fusing low-level features with high-level ones may result in redundancy and inconsistency. To address this, we propose the SBA module, which selectively aggregate the boundary information and semantic information to depict more fine-grained contour of objects and the location of re-calibrate objects. 

Different from previous fusion methods, we design a novel Re-calibration attention unit (RAU) block that adaptively picks up mutual representations from tow inputs ($F^s,F^b$) before fusion. As given in Figure \ref{fig1}, the shallow - and deep-level information is fed into the two RAU blocks by different ways to make up for the missing spatial boundary information of the high-level semantic features and the missing semantic information of low-level features. Finally, the outputs of two RAU blocks are concatenated after a $3 \times 3$ convolution. This aggregation strategy realizes the robust combination of different features and refines the rough features. The RAU block function $PAU(\cdot,\cdot)$ process can be expressed as:
\begin{gather}
	T_1' = W_{\theta}(T_1), T_2' =  W_{\phi}(T_2) \\
	PAU(T_1,T_2)= T_1' \odot T_1 + T_2' \odot T_2 \odot (\circleddash (T_1')) + T_1,
\end{gather}
where $T_1,T_2$ are the input features, two linear mapping and sigmoid functions $W_{\theta}(\cdot),W_{\phi}(\cdot)$ are applied to the input features to reduce the channel dimension to 32 and obtain feature maps $T_1'$ and $T_2'$. $\odot$ is Point-wise multiplication. $\circleddash(\cdot)$ is the reverse operation by subtracting the feature $T_1'$, refining the imprecise and coarse estimation into an accurate and complete prediction map \cite{fan2020pranet}. We take a convolutional operation with a kernel size of 1 $\times$ 1 as the linear mapping process. As a result, the process of SBA can be formulated as:\begin{gather}
    Z = C_{3 \times 3}({\rm Concat} (PAU(F^s,F^b), PAU(F^b,F^s))),
\end{gather}
where $C_{3 \times 3}(\cdot)$ is a $3 \times 3$ convolution with a batch normalization and a ReLU activation layer. $F^s \in \mathbb{R}^{\frac{H}{8} \times \frac{W}{8} \times 32}$ contains deep-level semantic information after fusing the third and fourth layers from the encoder, $F^b \in \mathbb{R}^{\frac{H}{4} \times \frac{W}{4} \times 32}$ is the first layer with rich boundary details from the backbone. $Concat(\cdot)$ is the concatenation operation along the channel dimension. $Z \in \mathbb{R}^{\frac{H}{4} \times \frac{W}{4} \times 32}$ is the output of the SBA module.

\begin{figure*}[t]
    \centerline{\includegraphics[width=\linewidth]{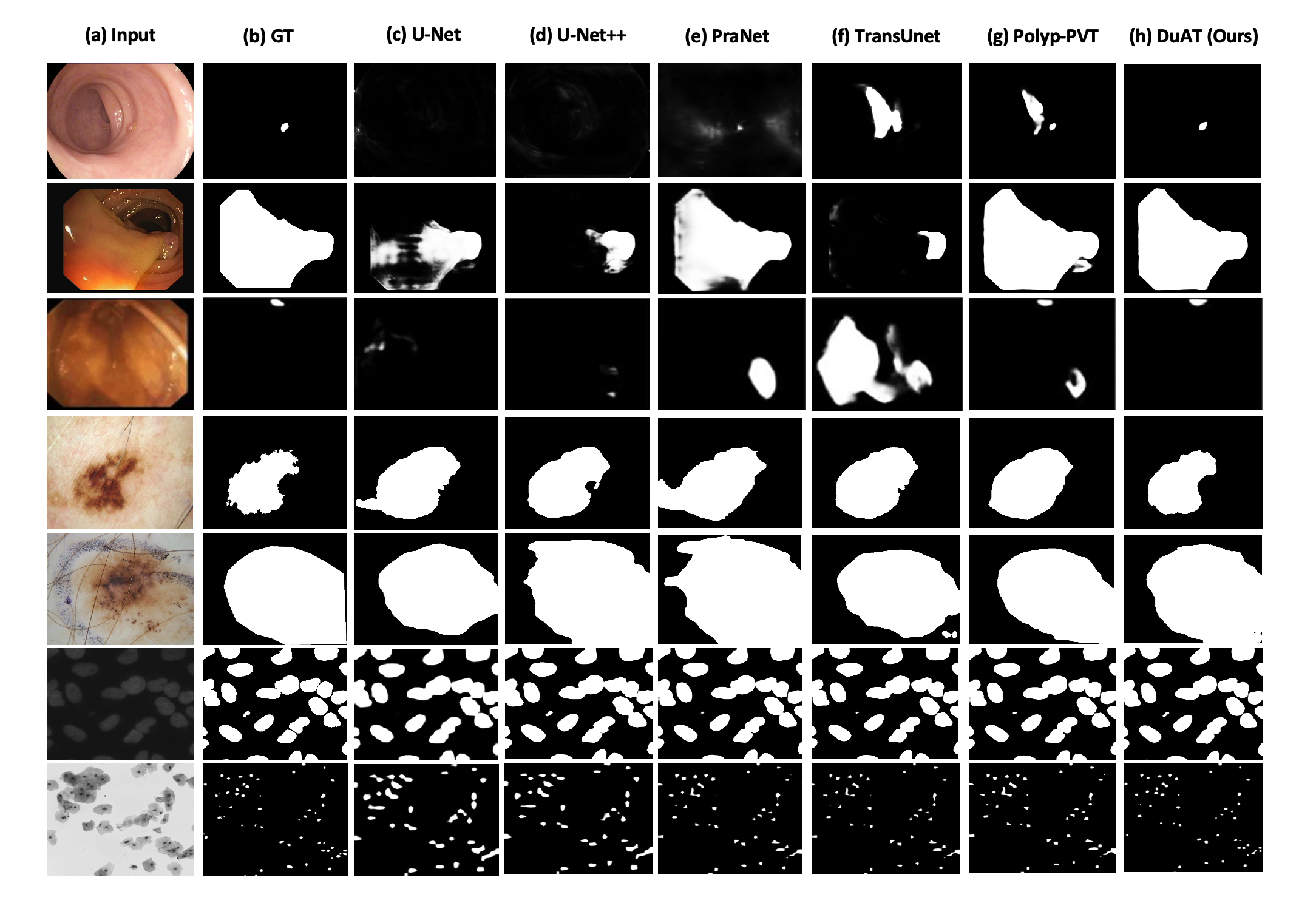}}
    \caption{Qualitative results of different methods. (a) Inputs images, (b) GT, which stands for the ground truths, (h) semantic segmentation maps produced by our method, (c) U-Net \cite{Unet}, (d) U-Net++ \cite{zhou2018unet++}, (e) PraNet \cite{fan2020pranet}, (f) TransUnet \cite{chen2021transunet}, (g) Polyp-PVT \cite{dong2021polyp}.}
    \label{fig4}
\end{figure*}

\subsection{Global-to-Local Spatial Aggregation} 
The attention mechanism strengthens the information related to the optimization goal and suppresses irrelevant information. In order to capture both global and local spatial features, we propose the GLSA module, which fuses the results of two separate local and global attention units. As demonstrated in Figure \ref{fig2}, this dual-stream design effectively preserves both local and non-local modeling capabilities. Moreover, we use separating channels to balance the accuracy and computational resources. Specifically, the feature map \{${\mathcal{F}_{i}|i \in (2,3,4) }$\} with 64 channels is split evenly into two feature map groups $\mathcal{F}_{i}^1, \mathcal{F}_{i}^2 (i \in (2,3,4))$ and  separately fed into Global Spatial attention (GSA) module and Local Spatial attention (LSA) module. The outputs of those two attention units are finally concatenated following by a 1 $\times$ 1 convolution layer. We formulate such a process as 
 

\begin{gather}
    \mathcal{F}_{i}^1, \mathcal{F}_{i}^2 = \rm Split(\mathcal{F}_{i}) \\
    \mathcal{F}_{i}^{'} = C_{1\times1}({\rm Concat}(G_{sa}(\mathcal{F}_{i}^1), L_{sa}(\mathcal{F}_{i}^2))).
\end{gather}

\begin{table*}[t]
    \centering
    \caption{Quantitative comparison of different methods on Kvasir, ClinicDB, ISIC-2018 and 2018-DSB datasets (seen datasets) to validate our model's learning ability. $\uparrow$ denotes higher the better and $\downarrow $ denotes lower the better.}
    \setlength{\tabcolsep}{1mm}{
    \begin{tabular}{c||ccc|ccc|ccc|ccc}
    \hline
    \multirow{2}{*}{Methods}   & \multicolumn{3}{c|}{Kvasir} & \multicolumn{3}{c|}{ClinicDB}   & \multicolumn{3}{c|}{ISIC-2018} & \multicolumn{3}{c}{2018-DSB}   \\ \cline{2-13}
    & mDice$\uparrow$  & mIou$\uparrow$  & MAE$\downarrow $  & mDice$\uparrow$  & mIou$\uparrow$ & MAE$\downarrow $   & mDice$\uparrow$   & mIou$\uparrow$  & MAE$\downarrow $  & mDice$\uparrow$   & mIou$\uparrow$  & MAE$\downarrow $ \\ \hline
    \makecell[c]{U-Net \cite{Unet}}                 & 0.818   & 0.746   & 0.055    & 0.823          & 0.755      & 0.019         & 0.855          & 0.785         & 0.045   & 0.908  & 0.831 & 0.040 \\
    \makecell[c]{UNet++ \cite{jha2019resunet++}}    & 0.821   & 0.743   & 0.048    & 0.794          & 0.729      & 0.022         & 0.809          & 0.729         & 0.041    & 0.911          & 0.837        & 0.039   \\
    \makecell[c]{PraNet \cite{fan2020pranet}}       & 0.898   & 0.840   & 0.030    & 0.899          & 0.849      & 0.009         & 0.875          & 0.787          & 0.037   & 0.912          & 0.838          & 0.036  \\
    \makecell[c]{CaraNet \cite{lou2021caranet}}   & 0.918   & 0.865      & 0.023    & 0.936  & 0.887 & 0.007    & 0.870  & 0.782          & 0.038   & 0.910   & 0.835   & 0.037 \\             
    \makecell[c]{TransUNet \cite{chen2021transunet}}    & 0.913   & 0.857           & 0.028          & 0.935          & 0.887          & 0.008              & 0.880          & 0.809          & 0.036   & 0.915          & 0.845       & 0.033                 \\ 
    \makecell[c]{TransFuse \cite{zhang2021transfuse}}   & 0.920   & 0.870           & 0.023              & 0.942          & 0.897          & 0.007              & 0.901          & 0.840          & 0.035  & 0.916          & 0.855    & 0.033               \\ 
    \makecell[c]{UCTransNet \cite{wang2022uctransnet}}   & 0.918   & 0.860           & 0.023              & 0.933          & 0.860          & 0.008              & 0.905     & 0.83   & 0.035  & 0.911          & 0.835    & 0.035               \\ 
    
    \makecell[c]{Polyp-PVT \cite{dong2021polyp}}        & 0.917    & 0.864          & 0.023          & 0.937          & 0.889         & 0.006          & 0.913          & 0.852          & 0.032     & 0.917         & 0.859       & 0.030        \\ \hline\hline

     \textbf{DuAT (Ours)}      & \textbf{0.924} & \textbf{0.876} & \textbf{0.023}   & \textbf{0.948} & \textbf{0.906} & \textbf{0.006}    & \textbf{0.923} & \textbf{0.867} & \textbf{0.029}  & \textbf{0.926} & \textbf{0.870} & \textbf{0.027}\\  \hline

    \end{tabular}}
    \label{table2}
\end{table*}

\begin{table*}[h]
    \centering
    \caption{Quantitative comparison of different methods on ColonDB, ETIS and EndoScene datasets (unseen datasets) to validate our model's generalization capability. $\uparrow$ denotes higher the better and $\downarrow $ denotes lower the better.}
    \setlength{\tabcolsep}{1mm}{
    \begin{tabular}{c||ccc|ccc|ccc}
    \hline
    \multirow{2}{*}{Methods}   & \multicolumn{3}{c|}{ColonDB} & \multicolumn{3}{c|}{ETIS}   & \multicolumn{3}{c}{EndoScene}   \\ \cline{2-10}
                  & mDice$\uparrow$    & mIou$\uparrow$           & MAE$\downarrow $         & mDice$\uparrow$    & mIou$\uparrow$           & MAE$\downarrow $          & mDice$\uparrow$    & mIou$\uparrow$           & MAE$\downarrow $                    \\ \hline
    \makecell[c]{U-Net \cite{Unet}}                    & 0.512   & 0.444          & 0.061          & 0.398          & 0.335          & 0.036          & 0.710          & 0.627          & 0.022                    \\
    \makecell[c]{UNet++ \cite{jha2019resunet++}}      & 0.483   & 0.410          & 0.064          & 0.401          & 0.344          & 0.035          & 0.707          & 0.624          & 0.018                    \\
    \makecell[c]{PraNet \cite{fan2020pranet}}          & 0.712   & 0.640          & 0.043          & 0.628          & 0.567          & 0.031          & 0.851          & 0.797          & 0.010                    \\
    \makecell[c]{CaraNet \cite{lou2021caranet}}   & 0.773   & 0.689     & 0.042    & 0.747          & 0.672        & 0.017             & 0.903         & 0.838         & 0.007   \\
    \makecell[c]{TransUNet \cite{chen2021transunet}}   & 0.781   & 0.699          & 0.036          & 0.731          & 0.824          & 0.021              & 0.893          & 0.660          &  0.009                   \\ 
    \makecell[c]{TransFuse \cite{zhang2021transfuse}}  & 0.781   & 0.706          & 0.035              & 0.737        & 0.826          & 0.020              & 0.894          & 0.654          & 0.009                    \\ 
    \makecell[c]{SSformer \cite{wang2022stepwise}}  & 0.772   & 0.697          & 0.036              & 0.767        & 0.698          & 0.016              & 0.887          & 0.821  & 0.007 \\
    \makecell[c]{Polyp-PVT \cite{dong2021polyp}}       & 0.808    & 0.727         & 0.031          & 0.787          & 0.706          & 0.013          & 0.900          & 0.833          & 0.007         \\ \hline\hline

    \makecell[c]{\textbf{DuAT (Ours)}}     & \textbf{0.819} & \textbf{0.737} & \textbf{0.026} & \textbf{0.822} & \textbf{0.746} & \textbf{0.013} &\textbf{0.901} & \textbf{0.840} & \textbf{0.005}  \\ \hline

    \end{tabular}}
    \label{table1}
\end{table*}

\noindent where $ G_{sa} $ denotes the global spatial attention and $L_{sa}$ denotes the local spatial attention. $\mathcal{F}_{i}^{'} \in \mathbb{R}^{\frac{H}{8} \times \frac{W}{8} \times 32}$ is the output features.
\noindent We will introduce LSA and GSA module in detail in the following.

$\left( 1 \right)$ GSA module: The GSA emphasizes the long-range relationship of each pixel in the spatial space and can be used as a supplement to local spatial attention. Many efforts \cite{bello2019attention}, \cite{cao2019gcnet} claim that the long-range interaction can make the feature more powerful. Inspired by the manners of extracting long-range interaction in \cite{bello2019attention}, we simply generate global spatial attention map ($G_{sa} \in \mathbb{R}^{\frac{H}{8}  \times\frac{W}{8} \times 32}$) and $\mathcal{F}_{i}^1$ as input as following: \begin{gather}
    Att_{G}(\mathcal{F}_{i}^1) = Softmax(Transpose(C_{1 \times 1} (\mathcal{F}_{i}^1))),\\
    G_{sa}(\mathcal{F}_{i}^1)= MLP(Att_{G}(\mathcal{F}_{i}^1)\otimes \mathcal{F}_{i}^1) + \mathcal{F}_{i}^1.
\end{gather}
where $Att_{G}(\cdot)$ is the attention operation, $C_{1 \times 1}$ means $1 \times 1$ convolution. $\otimes $ denotes matrix multiplication. $MLP(\cdot)$ consists of two fully-connection layers with a ReLU non-linearity and normalization layer. The first layer of MLP transforms its input to a higher-dimensional space which the expansion ratio is two, while the second layer restores the dimension to be the same as the input. 

$\left( 2 \right)$ LSA module: The LSA module extracts the local features of the region of interest effectively in the spatial dimension of the given feature map, such as small objects. In short, we compute local spatial attention response ($ L_{sa} \in \mathbb{R}^{\frac{H}{8}  \times\frac{W}{8} \times 32}$) and $\mathcal{F}_{i}^2$ as input as follow:


\begin{gather}
    Att_{L}(\mathcal{F}_{i}^2) = \sigma(C_{1 \times 1}(\mathcal{F}_c(\mathcal{F}_{i}^2))+\mathcal{F}_{i}^2)),\\
    L_{sa} = Att_{L}(\mathcal{F}_{i}^2)\odot \mathcal{F}_{i}^2  + \mathcal{F}_{i}^2.
\end{gather}

\noindent where $\mathcal{F}_c(\cdot)$ denotes cascading three $1\times1 $ convolution layers and $3\times3$ depth-wise convolution layers. The number of channels is adjusted to 32 in the $\mathcal{F}_c$. $Att_{L}(\cdot)$ is the local attention operation, $\sigma(\cdot)$ is the sigmoid function, $\odot$ is point-wise multiplication. This structural design can efficiently aggregate local spatial information using fewer parameters.

\subsection{Loss function} \cite{qin2019basnet, wei2020f3net} report that combining multiple loss functions with adaptive weights at different levels can improve the performance of the network with better convergence speed. Therefore, we use binary cross-entropy loss ($\mathcal{L}^{\omega}_{BCE}(\cdot)$) and the weighted IoU loss ( $\mathcal{L}^{\omega}_{Iou}(\cdot)$) for supervision. Our loss function is formulated in Eq.\ref{eq11}, where $S$ is the two side-outputs ($i,e., S_1,S_2$) and $G$ is the ground truth, respectively. $\lambda_1$ and $\lambda_2$ are the weighting coefficients. 
\begin{gather}
    \mathcal{L}(S,G) = \lambda_1 \mathcal{L}^{\omega}_{IoU}(S,G)+ \lambda_2 \mathcal{L}^{\omega}_{BCE}(S,G)
\label{eq11}
\end{gather}
\noindent Therefore, the total loss $\mathcal{L}_{total}$ for the proposed DuAT can be formulated as:
\begin{gather}
     \mathcal{L}_{total} = \mathcal{L}(S_1,G) +\mathcal{L}(S_2,G).
\label{eq12}
\end{gather}


\section{Experiments}

\subsection{Datasets}

In the experiment, we evaluate our proposed model on three different kinds of medical image sets: colonoscopy (Colon) images, dermoscopic (Derm) images, and microscopy (Micro) images, so as to assess the learning ability and generalization capability of our model. The detailed statistics of each dataset are shown in Table \ref{datset}.



\noindent \textbf{Colonoscopy polyp images}: Experiments are conducted on five polyp segmentation datasets (ETIS \cite{vazquez2017benchmark}, CVC-ClinicDB (ClinicDB) \cite{silva2014toward}, CVC-ColonDB (ColonDB) \cite{bernal2015wm}, EndoScene-CVC300 (EndoScene) \cite{tajbakhsh2015automated}, Kvasir-SEG (Kvasir) \cite{jha2020kvasir}). We follow the same training/testing protocols in \cite{dong2021polyp,fan2020pranet}, i.e., the images from the Kvasir and ClinicDB are randomly split into 80\% for training, 10\% for validation, and 10\% for testing (seen data). And test on the out-of-distribution datasets which are ColonDB with 380 images, EndoScene with 60 images and ETIS with 196 images (unseen data). Since the resolutions of images are not uniform, we resize them to 352$\times$352 resolution.

\noindent \textbf{ISIC-2018 Dataset}: The dataset comes from ISIC-2018 challenge \cite{codella2019skin} \cite{tschandl2018ham10000} and is useful for skin lesion analysis. It includes 2596 images and the corresponding annotations, which are resized to $512 \times 384 $ resolution. The images are randomly split into 80\% for training, 10\% for validation, and 10\% for testing. 

\noindent \textbf{2018 Data Science Bowl (2018-DSB)}: The dataset comes from 2018 Data Science Bowl challenge \cite{caicedo2019nucleus} and is used to find the nuclei in divergent images, including 670 images and the corresponding annotations, which are resized to $256 \times 256 $ resolution. The images are randomly split into 80\% for training, 10\% for validation, and 10\% for testing.



\begin{table}[t]
\scriptsize
\centering
\caption{Statistics on polyp, ISIC-2018 and 2018-DSB datasets.}
\begin{tabular}{l|c|c|c|c|c|c} \hline 
\toprule[1pt]
Dataset & Imaging & Images  & Shape & Train  & Valid  & Test \\ \hline \hline
Kvasir & Colon & 1000  & Variable  & 800  & 100 & 100 \\ \hline
ClinicDB & Colon & 612  & 384 $\times$ 288  & 488 & 62 & 62  \\  \hline
ColonDB & Colon & 380  & 574 $\times$ 500  & -   & - & 380 \\ \hline
ETIS & Colon & 196 & 1225 $\times$ 966 & -  & - & 196 \\ \hline
EndoScene & Colon &  60 & 574 $\times$ 500  & -   & - & 60 \\ \hline
ISIC-2018 & Derm & 2596  & 512 $\times$ 384 & 2078   & 259 & 259 \\ \hline  
2018-DSB & Micro & 670  & 256 $\times$ 256  & 536 & 67 & 67 \\ \hline 

\bottomrule[1pt]
\end{tabular}
\label{datset}
\end{table}

\begin{figure*}[t]
    \centerline{\includegraphics[width=\linewidth]{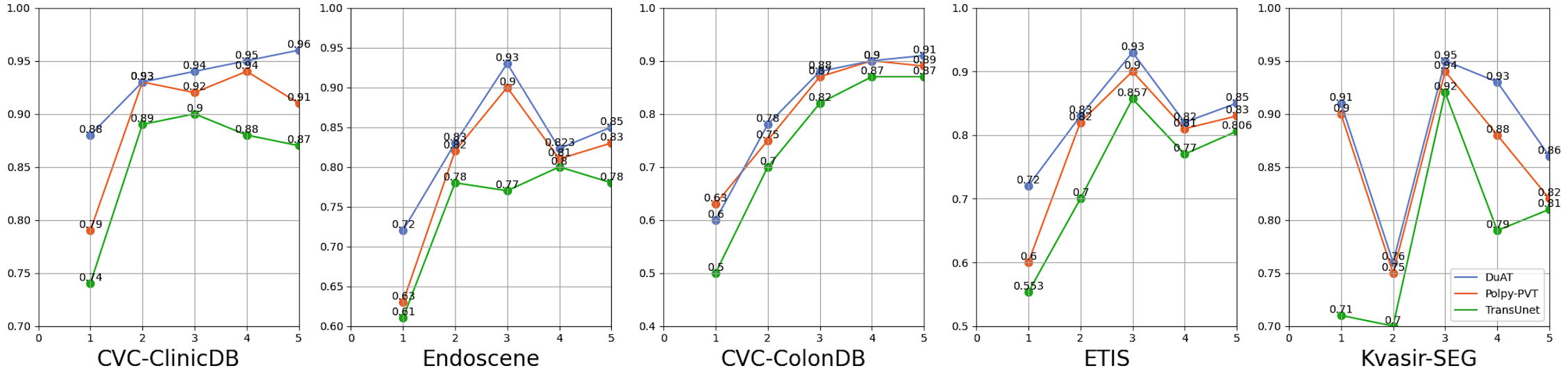}}
    \caption{Performance vs. Size on the five polyp datasets. The x-axis is the proportion size \% of polyp and y-axis is the averaged mDice coefficient. Blue is for our \textcolor{blue}{DuAT}, orange is for the \textcolor{orange}{Polyp-PVT}, green is for the \textcolor{green}{TransUnet}.}
    \label{fig5}
\end{figure*}

\subsection{Evaluation Metrics and Implementation Details} 

\noindent \textbf{Evaluation Metrics.} We employ three widely-used metrics i.e., mean Dice (mDice), mean IoU (mIoU) and mean absolute error (MAE) to evaluate the model performances. Mean Dice and IoU are the most commonly used metrics and mainly emphasise the internal consistency of segmentation results. MAE is used to evaluate the pixel-level accuracy representing the average absolute error between the prediction and true values. 

\noindent \textbf{Implementation Details.} We use rotation and horizontal flip for data augmentation. Considering the differences in the sizes and color of each polyp image, we adopt a multi-scale training \cite{fan2020pranet,huang2021hardnet} and the color exchange \cite{wei2021shallow}. The network is trained end-to-end by AdamW \cite{loshchilov2017decoupled} optimizer. The learning rate is set to 1e-4 and the weight decay is adjusted to 1e-4 too. The batch size is set at 16. We use PyTorch framework for implementation with an NVIDIA RTX 3090 GPU. We will provide the source code after the paper is published.

\subsection{Results}

\textbf{Learning Ability}. We first evaluate our proposed DuAT model for its segmentation performance on seen datasets. As summarized in Table \ref{table2}, our model is compared to six recently published models: U-Net \cite{Unet}, UNet++ \cite{jha2019resunet++}, PraNet \cite{fan2020pranet}, CaraNet \cite{lou2021caranet}, TransUNet \cite{chen2021transunet}, TransFuse \cite{zhang2021transfuse}, TransFuse \cite{zhang2021transfuse} and Polyp-PVT \cite{dong2021polyp} . It can be observed that our DuAT model outperforms all other models, and achieving 0.924 and 0.948 mean Dice scores on Kvasir and ClinicDB segmentation respectively. For ISIC-2018 dataset, our DuAT model achieves a 1.0\% improvement in terms of mDice and 1.5\% of mIoU over SOTA method. For 2018-DSB, DuAT achieves a mIoU of 0.87, mDice of 0.926 and 0.027 of MAE, which are 1.1\%, 1.0\%, 0.03\% higher than the best performing Polyp-PVT. These results demonstrate that our model can effectively segment polyps.


\textbf{Generalization Capabilities.} We further evaluate the generalisation capability of our model on unseen datasets (ETIS, ColonDB, EndoScene). These three datasets have their own specific challenges and properties. For example, ColonDB is a small-scale database that contains 380 images from 15 short colonoscopy sequences. ETIS consists of 196 polyp images for early diagnosis of colorectal cancer. EndoScene is a re-annotated branch with associated polyp and background (mucosa and lumen). As seen in Table \ref{table1}, our model outperforms the existing medical segmentation baselines on all unseen datasets for all metrics. Moreover, our DuAT is able to achieve an average dice of 82.2 \% on the most challenging ETIS dataset, 3.5\% higher than Polyp-PVT.

\textbf{Visual Results.} We also demonstrate qualitative the performance of our model on five benchmarks, as given in Figure \ref{fig4}. On ETIS (the first and second row), DuAT is able to accurately capture the target object's boundary and detect a small polyp while other methods fail to detect. On ISIC-2018 (third row), all methods are able to segment the lesion skin, but our method show the most similar results compared to the ground truth. On 2018-DSB (the fourth row), we can observe that our DuAT is better able to capture the presence of nuclei and obtain better segmentation predictions. More qualitative results can be found in the supplementary material.

\textbf{Computational Efficiency.} Table \ref{table122} presents the number of parameters and floating-point operations for different methods. As our proposed DuAT and Polyp-PVT \cite{dong2021polyp} adopt the same backbone, they have similar model size (Params). DuAT uses 24.92M of parameter and 9.88G of FLOPs, which is more lightweight and compact than CNN-based neural network and Transformer-based methods.

\begin{table}
  \begin{center}
    {\small{
\caption{Number of Parameters and FLOPs on a 352 × 352 input image. Note that ``N/A" means the result is not available.}
\begin{tabular}{llll}
\toprule
Method & Type & Params(M) & FLOPs(G) \\
\midrule
U-Net \cite{Unet} & CNN & 43.93 & 43.47\\
UNet++ \cite{jha2019resunet++} & CNN & 9.04 & 64.03\\
PraNet \cite{fan2020pranet} & CNN & 30.50 & 13.01\\
CaraNet \cite{lou2021caranet} & CNN & 44.59 & 21.65 \\
TransUNet \cite{chen2021transunet} & Transformer & 93.19 & 60.84\\
SSformer \cite{wang2022stepwise} & Transformer & 26.70 & 28.07\\
TransFuse \cite{zhang2021transfuse} & Transformer & N/A & N/A\\
Polyp-PVT \cite{dong2021polyp} & Transformer & 25.08 & 10.00\\
\textbf{Ours} & Transformer & 24.92 & 9.88 \\
\bottomrule
\label{table122}
\end{tabular}
}}
\end{center}
\end{table}

       

\begin{table*}[h]
    \centering
    \caption{Ablation study for DuAT on the Kvasir, ETIS and ISIC-2018 datasets. $\uparrow$ denotes higher the better and $\downarrow $ denotes lower the better.}
    \setlength{\tabcolsep}{1mm}
    \begin{tabular}{c||ccc|ccc|ccc}
    \hline
    \multirow{2}*{Methods}                     & \multicolumn{3}{c|}{Kvasir-SEG (seen)}    & \multicolumn{3}{c|}{ETIS (unseen)}  & \multicolumn{3}{c}{ISIC-2018 (seen)} \\ \cline{2-10}
    & mDice$\uparrow$    & mIou$\uparrow$           & MAE$\downarrow $            & mDice$\uparrow$    & mIou$\uparrow$           & MAE$\downarrow $   & mDice$\uparrow$    & mIou$\uparrow$           & MAE$\downarrow $      \\ \hline
    \makecell[c]{Baseline}                     & 0.910 & 0.856  & 0.030         & 0.759 & 0.668  & 0.035  & 0.877 & 0.783  & 0.040   \\ 
    \makecell[c]{\textit{+ GSA}}               & 0.912 & 0.860  & 0.029         & 0.772 & 0.675  & 0.030  & 0.887 & 0.803  & 0.038   \\ 
    \makecell[c]{\textit{+ LSA}}               & 0.916 & 0.863  & 0.028         & 0.785 & 0.690  & 0.027  & 0.900 & 0.839  & 0.035    \\
    \makecell[c]{\textit{+ GSA + LSA (Serial)}} & 0.914 & 0.863  & 0.028         & 0.786 & 0.695  & 0.025  & 0.909 & 0.845  & 0.034      \\
    \makecell[c]{\textit{+ LSA + GSA (Serial)}}          & 0.910 & 0.860  & 0.029         & 0.799 & 0.713  & 0.021  & 0.910 & 0.852  & 0.033        \\ 
    \makecell[c]{\textit{+GLSA}}  & 0.917 & 0.864  & 0.025         & 0.814 & 0.723  & 0.016  & 0.916 & 0.816  & 0.031 \\ \hline
    \makecell[c]{\textit{w/o SBA}}                     & 0.917 & 0.864  & 0.025         & 0.814 & 0.723  & 0.016  & 0.916 & 0.816  & 0.031        \\  
    \makecell[c]{\textit{w/o GLSA}}                      & 0.915 & 0.863  & 0.026         & 0.790 & 0.696  & 0.023  & 0.901 & 0.800  & 0.033       \\   \hline \hline
   \makecell[c]{\textit{\textbf{SBA + GLSA (Ours)}}}          & \textbf{0.924} & \textbf{0.876}  & \textbf{0.023}         & \textbf{0.822} & \textbf{0.746}  & \textbf{0.013}  & \textbf{0.923} & \textbf{0.867}  & \textbf{0.029} \\ \hline
    \end{tabular}
    \label{table3}
\end{table*}

\textbf{Small Object Segmentation Analysis.} To demonstrate the detection ability of our model for small objects, the ratio of the number of pixels in the object to the number of pixels in the entire image is used to account for the size of the object. We then evaluate the performance of the segmentation model based on the size of the object. We set the area with a proportion less than 5\%. For the segmentation model, we first obtain the mean Dice coefficient of the five polyp datasets. Similar to computing the histogram, we calculate the average of mean Dice of test data whose size values fall into each interval. For the small object segmentation analysis, we compare our DuAT with Polyp-PVT and TransUnet, and the results are given in Figure \ref{fig5}. The overall accuracy of DuAT is higher than TransUnet \cite{chen2021transunet} and Polyp-PVT \cite{wang2022pvt} on samples with small size polyps.

\subsection{Ablation Study}
We further conduct ablation study to demonstrate the necessity and effectiveness of each component of our proposed model on three datasets, and we choose mDice, mIoU and MAE for evaluation.

\textbf{Effectiveness of SBA and GLSA.} We conduct an experiment to evaluate DuAT without SBA module ``(w/o SBA)". The performance without the SBA drops sharply on all three datasets are shown in Table \ref{table3}. In particular, the mDice is reduced from 0.822 to 0.814 on ETIS. Moreover, we further investigate the contribution of the Global-to-Local Spatial Aggregation by removing it from the overall DuAT and replacing it with convolution operation with a kernel size of 3, which is denoted as “(w/o GLSA)”. The performance of the complete DuAT shows an improvement of 2.2 \% and 6.7\% in terms of mDice and mIoU respectively on ISIC-2018. After using the two modules (SBA + GLSA), the model’s performance is improved again. These results demonstrate that these modules enable our model to distinguish polyp and lesion tissues effectively.

\begin{figure}[h]
    \footnotesize
    \centerline{\includegraphics[width=\linewidth]{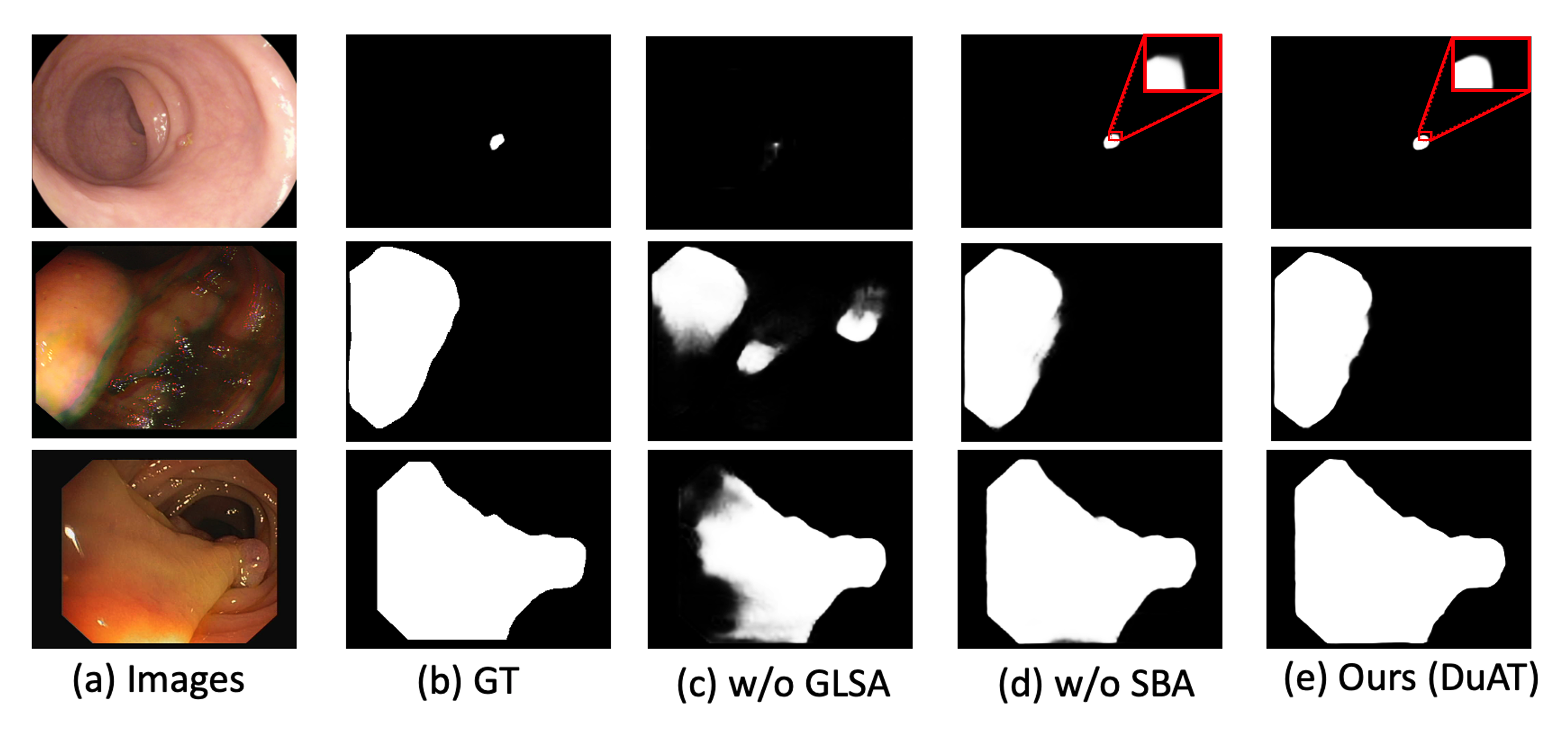}}
    \caption{The effectiveness of each component.}
    \label{fig8}
\end{figure}

The visual results are given in Figure \ref{fig8}. We observe that the SBA module facilitates the fine-grained of ambiguous boundaries and the GLSA module greatly improves the accuracies of small object detection and target object location.

\textbf{Arrangements of GSA and LSA.} \textit{GSA} and \textit{LSA} represent global spatial attention module and local spatial attention module respectively. We further study the effectiveness and different arrangements of \textit{GSA} and \textit{LSA}. The results tested on Kavsir, ETIS, and ISIC-2018 datasets are shown in Table \ref{table3} (the second and sixth row), and all the methods are using the same backbone PVTv2 \cite{wang2022pvt}. \textit{GSA + LSA(Serial)} means first performing \textit{GSA} then \textit{LSA}, while \textit{LSA + GSA(Serial)} is the opposite. Overall, all improve the baseline and our GLSA group achieves more accuracy and reliable results. The GLSA module outperforms the \textit{GSA}, \textit{LSA}, \textit{GSA + LSA (Serial)},\textit{ LSA + GSA (Serial)} by 4.2\%, 2.9\%, 2.8\%, 1.5\% in term of mean \textit{Dice} on the ETIS dataset.

\section{Conclusions}

In this work, we propose DuAT to address the issues related to medical image segmentation. Two components, the \textit{GLSA} and \textit{SBA} are proposed. Specifically, the \textit{GLSA} module extracts the global and local spatial features from the encoder and is beneficial for locating the large and small objects. The \textit{SBA} module alleviates the unclear boundary of high-level features and further improves its performance. As a result, DuAT can achieve strong learning, generalization ability, and lightweight segmentation efficiency. Both qualitative and quantitative results demonstrate the superiority of our DuAT over other competing methods. We hope that this research will inspire more ideas to solve the medical image segmentation task and we will extend the proposed model to tackle 3D medical image segmentation task in the future work.

\section*{Acknowledgments} This work was supported by the Key Program Special Fund in XJTLU (KSF-A-22).

{\small
\bibliographystyle{ieee_fullname}
\bibliography{egbib}

\begin{thebibliography}{10}\itemsep=-1pt

\bibitem{beal2020toward}
Josh Beal, Eric Kim, Eric Tzeng, Dong~Huk Park, Andrew Zhai, and Dmitry
  Kislyuk.
\newblock Toward transformer-based object detection.
\newblock {\em arXiv preprint arXiv:2012.09958}, 2020.

\bibitem{bello2019attention}
Irwan Bello, Barret Zoph, Ashish Vaswani, Jonathon Shlens, and Quoc~V Le.
\newblock Attention augmented convolutional networks.
\newblock In {\em Proceedings of the IEEE/CVF international conference on
  computer vision}, pages 3286--3295, 2019.

\bibitem{bernal2015wm}
Jorge Bernal, F~Javier S{\'a}nchez, Gloria Fern{\'a}ndez-Esparrach, Debora Gil,
  Cristina Rodr{\'\i}guez, and Fernando Vilari{\~n}o.
\newblock Wm-dova maps for accurate polyp highlighting in colonoscopy:
  Validation vs. saliency maps from physicians.
\newblock {\em Computerized Medical Imaging and Graphics}, 43:99--111, 2015.

\bibitem{bertasius2016semantic}
Gedas Bertasius, Jianbo Shi, and Lorenzo Torresani.
\newblock Semantic segmentation with boundary neural fields.
\newblock In {\em Proceedings of the IEEE conference on computer vision and
  pattern recognition}, pages 3602--3610, 2016.

\bibitem{bhojanapalli2021understanding}
Srinadh Bhojanapalli, Ayan Chakrabarti, Daniel Glasner, Daliang Li, Thomas
  Unterthiner, and Andreas Veit.
\newblock Understanding robustness of transformers for image classification.
\newblock In {\em Proceedings of the IEEE/CVF International Conference on
  Computer Vision}, pages 10231--10241, 2021.

\bibitem{caicedo2019nucleus}
Juan~C Caicedo, Allen Goodman, Kyle~W Karhohs, Beth~A Cimini, Jeanelle
  Ackerman, Marzieh Haghighi, CherKeng Heng, Tim Becker, Minh Doan, Claire
  McQuin, et~al.
\newblock Nucleus segmentation across imaging experiments: the 2018 data
  science bowl.
\newblock {\em Nature methods}, 16(12):1247--1253, 2019.

\bibitem{cao2019gcnet}
Yue Cao, Jiarui Xu, Stephen Lin, Fangyun Wei, and Han Hu.
\newblock Gcnet: Non-local networks meet squeeze-excitation networks and
  beyond.
\newblock In {\em Proceedings of the IEEE/CVF International Conference on
  Computer Vision Workshops}, pages 0--0, 2019.

\bibitem{carion2020end}
Nicolas Carion, Francisco Massa, Gabriel Synnaeve, Nicolas Usunier, Alexander
  Kirillov, and Sergey Zagoruyko.
\newblock End-to-end object detection with transformers.
\newblock In {\em European conference on computer vision}, pages 213--229.
  Springer, 2020.

\bibitem{chen2021pre}
Hanting Chen, Yunhe Wang, Tianyu Guo, Chang Xu, Yiping Deng, Zhenhua Liu, Siwei
  Ma, Chunjing Xu, Chao Xu, and Wen Gao.
\newblock Pre-trained image processing transformer.
\newblock In {\em Proceedings of the IEEE/CVF Conference on Computer Vision and
  Pattern Recognition}, pages 12299--12310, 2021.

\bibitem{chen2021transunet}
Jieneng Chen, Yongyi Lu, Qihang Yu, Xiangde Luo, Ehsan Adeli, Yan Wang, Le Lu,
  Alan~L Yuille, and Yuyin Zhou.
\newblock Transunet: Transformers make strong encoders for medical image
  segmentation.
\newblock {\em arXiv preprint arXiv:2102.04306}, 2021.

\bibitem{chen2017deeplab}
Liang-Chieh Chen, George Papandreou, Iasonas Kokkinos, Kevin Murphy, and Alan~L
  Yuille.
\newblock Deeplab: Semantic image segmentation with deep convolutional nets,
  atrous convolution, and fully connected crfs.
\newblock {\em IEEE transactions on pattern analysis and machine intelligence},
  40(4):834--848, 2017.

\bibitem{chen2018encoder}
Liang-Chieh Chen, Yukun Zhu, George Papandreou, Florian Schroff, and Hartwig
  Adam.
\newblock Encoder-decoder with atrous separable convolution for semantic image
  segmentation.
\newblock In {\em Proceedings of the European conference on computer vision
  (ECCV)}, pages 801--818, 2018.

\bibitem{codella2019skin}
Noel Codella, Veronica Rotemberg, Philipp Tschandl, M~Emre Celebi, Stephen
  Dusza, David Gutman, Brian Helba, Aadi Kalloo, Konstantinos Liopyris, Michael
  Marchetti, et~al.
\newblock Skin lesion analysis toward melanoma detection 2018: A challenge
  hosted by the international skin imaging collaboration (isic).
\newblock {\em arXiv preprint arXiv:1902.03368}, 2019.

\bibitem{dong2021polyp}
Bo Dong, Wenhai Wang, Deng-Ping Fan, Jinpeng Li, Huazhu Fu, and Ling Shao.
\newblock Polyp-pvt: Polyp segmentation with pyramid vision transformers.
\newblock {\em arXiv preprint arXiv:2108.06932}, 2021.

\bibitem{dosovitskiy2020image}
Alexey Dosovitskiy, Lucas Beyer, Alexander Kolesnikov, Dirk Weissenborn,
  Xiaohua Zhai, Thomas Unterthiner, Mostafa Dehghani, Matthias Minderer, Georg
  Heigold, Sylvain Gelly, et~al.
\newblock An image is worth 16x16 words: Transformers for image recognition at
  scale.
\newblock {\em arXiv preprint arXiv:2010.11929}, 2020.

\bibitem{fan2020pranet}
Deng-Ping Fan, Ge-Peng Ji, Tao Zhou, Geng Chen, Huazhu Fu, Jianbing Shen, and
  Ling Shao.
\newblock Pranet: Parallel reverse attention network for polyp segmentation.
\newblock In {\em International conference on medical image computing and
  computer-assisted intervention}, pages 263--273. Springer, 2020.

\bibitem{favoriti2016worldwide}
Pasqualino Favoriti, Gabriele Carbone, Marco Greco, Felice Pirozzi, Raffaele
  Emmanuele~Maria Pirozzi, and Francesco Corcione.
\newblock Worldwide burden of colorectal cancer: a review.
\newblock {\em Updates in surgery}, 68(1):7--11, 2016.

\bibitem{fu2018joint}
Huazhu Fu, Jun Cheng, Yanwu Xu, Damon Wing~Kee Wong, Jiang Liu, and Xiaochun
  Cao.
\newblock Joint optic disc and cup segmentation based on multi-label deep
  network and polar transformation.
\newblock {\em IEEE transactions on medical imaging}, 37(7):1597--1605, 2018.

\bibitem{han2021transformer}
Kai Han, An Xiao, Enhua Wu, Jianyuan Guo, Chunjing Xu, and Yunhe Wang.
\newblock Transformer in transformer.
\newblock {\em Advances in Neural Information Processing Systems}, 34, 2021.

\bibitem{huang2021hardnet}
Chien-Hsiang Huang, Hung-Yu Wu, and Youn-Long Lin.
\newblock Hardnet-mseg: a simple encoder-decoder polyp segmentation neural
  network that achieves over 0.9 mean dice and 86 fps.
\newblock {\em arXiv preprint arXiv:2101.07172}, 2021.

\bibitem{huang2019ccnet}
Zilong Huang, Xinggang Wang, Lichao Huang, Chang Huang, Yunchao Wei, and Wenyu
  Liu.
\newblock Ccnet: Criss-cross attention for semantic segmentation.
\newblock In {\em Proceedings of the IEEE/CVF International Conference on
  Computer Vision}, pages 603--612, 2019.

\bibitem{jha2020kvasir}
Debesh Jha, Pia~H Smedsrud, Michael~A Riegler, P{\aa}l Halvorsen, Thomas~de
  Lange, Dag Johansen, and H{\aa}vard~D Johansen.
\newblock Kvasir-seg: A segmented polyp dataset.
\newblock In {\em International Conference on Multimedia Modeling}, pages
  451--462. Springer, 2020.

\bibitem{jha2019resunet++}
Debesh Jha, Pia~H Smedsrud, Michael~A Riegler, Dag Johansen, Thomas De~Lange,
  P{\aa}l Halvorsen, and H{\aa}vard~D Johansen.
\newblock Resunet++: An advanced architecture for medical image segmentation.
\newblock In {\em 2019 IEEE International Symposium on Multimedia (ISM)}, pages
  225--2255. IEEE, 2019.

\bibitem{ji2022fast}
Ge-Peng Ji, Lei Zhu, Mingchen Zhuge, and Keren Fu.
\newblock Fast camouflaged object detection via edge-based reversible
  re-calibration network.
\newblock {\em Pattern Recognition}, 123:108414, 2022.

\bibitem{li2020improving}
Xiangtai Li, Xia Li, Li Zhang, Guangliang Cheng, Jianping Shi, Zhouchen Lin,
  Shaohua Tan, and Yunhai Tong.
\newblock Improving semantic segmentation via decoupled body and edge
  supervision.
\newblock In {\em European Conference on Computer Vision}, pages 435--452.
  Springer, 2020.

\bibitem{li2019global}
Xiangtai Li, Li Zhang, Ansheng You, Maoke Yang, Kuiyuan Yang, and Yunhai Tong.
\newblock Global aggregation then local distribution in fully convolutional
  networks.
\newblock {\em arXiv preprint arXiv:1909.07229}, 2019.

\bibitem{li2020gated}
Xiangtai Li, Houlong Zhao, Lei Han, Yunhai Tong, Shaohua Tan, and Kuiyuan Yang.
\newblock Gated fully fusion for semantic segmentation.
\newblock In {\em Proceedings of the AAAI conference on artificial
  intelligence}, volume~34, pages 11418--11425, 2020.

\bibitem{liu2021swin}
Ze Liu, Yutong Lin, Yue Cao, Han Hu, Yixuan Wei, Zheng Zhang, Stephen Lin, and
  Baining Guo.
\newblock Swin transformer: Hierarchical vision transformer using shifted
  windows.
\newblock In {\em Proceedings of the IEEE/CVF International Conference on
  Computer Vision}, pages 10012--10022, 2021.

\bibitem{loshchilov2017decoupled}
Ilya Loshchilov and Frank Hutter.
\newblock Decoupled weight decay regularization.
\newblock {\em arXiv preprint arXiv:1711.05101}, 2017.

\bibitem{lou2021caranet}
Ange Lou, Shuyue Guan, and Murray Loew.
\newblock Caranet: Context axial reverse attention network for segmentation of
  small medical objects.
\newblock {\em arXiv preprint arXiv:2108.07368}, 2021.

\bibitem{lu2019graph}
Yi Lu, Yaran Chen, Dongbin Zhao, and Jianxin Chen.
\newblock Graph-fcn for image semantic segmentation.
\newblock In {\em International symposium on neural networks}, pages 97--105.
  Springer, 2019.

\bibitem{ma2021boundary}
Haoxiang Ma, Hongyu Yang, and Di Huang.
\newblock Boundary guided context aggregation for semantic segmentation.
\newblock {\em arXiv preprint arXiv:2110.14587}, 2021.

\bibitem{mathur2020cancer}
Prashant Mathur, Krishnan Sathishkumar, Meesha Chaturvedi, Priyanka Das,
  Kondalli~Lakshminarayana Sudarshan, Stephen Santhappan, Vinodh Nallasamy,
  Anish John, Sandeep Narasimhan, Francis~Selvaraj Roselind, et~al.
\newblock Cancer statistics, 2020: report from national cancer registry
  programme, india.
\newblock {\em JCO Global Oncology}, 6:1063--1075, 2020.

\bibitem{qin2019basnet}
Xuebin Qin, Zichen Zhang, Chenyang Huang, Chao Gao, Masood Dehghan, and Martin
  Jagersand.
\newblock Basnet: Boundary-aware salient object detection.
\newblock In {\em Proceedings of the IEEE/CVF conference on computer vision and
  pattern recognition}, pages 7479--7489, 2019.

\bibitem{ranftl2021vision}
Ren{\'e} Ranftl, Alexey Bochkovskiy, and Vladlen Koltun.
\newblock Vision transformers for dense prediction.
\newblock In {\em Proceedings of the IEEE/CVF International Conference on
  Computer Vision}, pages 12179--12188, 2021.

\bibitem{Unet}
Olaf Ronneberger, Philipp Fischer, and Thomas Brox.
\newblock U-net: Convolutional networks for biomedical image segmentation.
\newblock In {\em International Conference on Medical image computing and
  computer-assisted intervention}, pages 234--241. Springer, 2015.

\bibitem{silva2014toward}
Juan Silva, Aymeric Histace, Olivier Romain, Xavier Dray, and Bertrand Granado.
\newblock Toward embedded detection of polyps in wce images for early diagnosis
  of colorectal cancer.
\newblock {\em International journal of computer assisted radiology and
  surgery}, 9(2):283--293, 2014.

\bibitem{tajbakhsh2015automated}
Nima Tajbakhsh, Suryakanth~R Gurudu, and Jianming Liang.
\newblock Automated polyp detection in colonoscopy videos using shape and
  context information.
\newblock {\em IEEE transactions on medical imaging}, 35(2):630--644, 2015.

\bibitem{takikawa2019gated}
Towaki Takikawa, David Acuna, Varun Jampani, and Sanja Fidler.
\newblock Gated-scnn: Gated shape cnns for semantic segmentation.
\newblock In {\em Proceedings of the IEEE/CVF international conference on
  computer vision}, pages 5229--5238, 2019.

\bibitem{touvron2021training}
Hugo Touvron, Matthieu Cord, Matthijs Douze, Francisco Massa, Alexandre
  Sablayrolles, and Herv{\'e} J{\'e}gou.
\newblock Training data-efficient image transformers \& distillation through
  attention.
\newblock In {\em International Conference on Machine Learning}, pages
  10347--10357. PMLR, 2021.

\bibitem{tschandl2018ham10000}
Philipp Tschandl, Cliff Rosendahl, and Harald Kittler.
\newblock The ham10000 dataset, a large collection of multi-source
  dermatoscopic images of common pigmented skin lesions.
\newblock {\em Scientific data}, 5(1):1--9, 2018.

\bibitem{vaswani2017attention}
Ashish Vaswani, Noam Shazeer, Niki Parmar, Jakob Uszkoreit, Llion Jones,
  Aidan~N Gomez, {\L}ukasz Kaiser, and Illia Polosukhin.
\newblock Attention is all you need.
\newblock {\em Advances in neural information processing systems}, 30, 2017.

\bibitem{vazquez2017benchmark}
David V{\'a}zquez, Jorge Bernal, F~Javier S{\'a}nchez, Gloria
  Fern{\'a}ndez-Esparrach, Antonio~M L{\'o}pez, Adriana Romero, Michal
  Drozdzal, and Aaron Courville.
\newblock A benchmark for endoluminal scene segmentation of colonoscopy images.
\newblock {\em Journal of healthcare engineering}, 2017, 2017.

\bibitem{wang2022uctransnet}
Haonan Wang, Peng Cao, Jiaqi Wang, and Osmar~R Zaiane.
\newblock Uctransnet: rethinking the skip connections in u-net from a
  channel-wise perspective with transformer.
\newblock In {\em Proceedings of the AAAI Conference on Artificial
  Intelligence}, volume~36, pages 2441--2449, 2022.

\bibitem{wang2022stepwise}
Jinfeng Wang, Qiming Huang, Feilong Tang, Jia Meng, Jionglong Su, and Sifan
  Song.
\newblock Stepwise feature fusion: Local guides global.
\newblock {\em arXiv preprint arXiv:2203.03635}, 2022.

\bibitem{wang2022pvt}
Wenhai Wang, Enze Xie, Xiang Li, Deng-Ping Fan, Kaitao Song, Ding Liang, Tong
  Lu, Ping Luo, and Ling Shao.
\newblock Pvt v2: Improved baselines with pyramid vision transformer.
\newblock {\em Computational Visual Media}, pages 1--10, 2022.

\bibitem{wang2018non}
Xiaolong Wang, Ross Girshick, Abhinav Gupta, and Kaiming He.
\newblock Non-local neural networks.
\newblock In {\em Proceedings of the IEEE conference on computer vision and
  pattern recognition}, pages 7794--7803, 2018.

\bibitem{wei2021shallow}
Jun Wei, Yiwen Hu, Ruimao Zhang, Zhen Li, S~Kevin Zhou, and Shuguang Cui.
\newblock Shallow attention network for polyp segmentation.
\newblock In {\em International Conference on Medical Image Computing and
  Computer-Assisted Intervention}, pages 699--708. Springer, 2021.

\bibitem{wei2020f3net}
Jun Wei, Shuhui Wang, and Qingming Huang.
\newblock F$^3$net: fusion, feedback and focus for salient object detection.
\newblock In {\em Proceedings of the AAAI Conference on Artificial
  Intelligence}, volume~34, pages 12321--12328, 2020.

\bibitem{xie2021segformer}
Enze Xie, Wenhai Wang, Zhiding Yu, Anima Anandkumar, Jose~M Alvarez, and Ping
  Luo.
\newblock Segformer: Simple and efficient design for semantic segmentation with
  transformers.
\newblock {\em Advances in Neural Information Processing Systems}, 34, 2021.

\bibitem{yu2015multi}
Fisher Yu and Vladlen Koltun.
\newblock Multi-scale context aggregation by dilated convolutions.
\newblock {\em arXiv preprint arXiv:1511.07122}, 2015.

\bibitem{zhang2019dual}
Li Zhang, Xiangtai Li, Anurag Arnab, Kuiyuan Yang, Yunhai Tong, and Philip~HS
  Torr.
\newblock Dual graph convolutional network for semantic segmentation.
\newblock {\em arXiv preprint arXiv:1909.06121}, 2019.

\bibitem{zhang2020dynamic}
Li Zhang, Dan Xu, Anurag Arnab, and Philip~HS Torr.
\newblock Dynamic graph message passing networks.
\newblock In {\em Proceedings of the IEEE/CVF Conference on Computer Vision and
  Pattern Recognition}, pages 3726--3735, 2020.

\bibitem{zhang2020adaptive}
Ruifei Zhang, Guanbin Li, Zhen Li, Shuguang Cui, Dahong Qian, and Yizhou Yu.
\newblock Adaptive context selection for polyp segmentation.
\newblock In {\em International Conference on Medical Image Computing and
  Computer-Assisted Intervention}, pages 253--262. Springer, 2020.

\bibitem{zhang2021transfuse}
Yundong Zhang, Huiye Liu, and Qiang Hu.
\newblock Transfuse: Fusing transformers and cnns for medical image
  segmentation.
\newblock In {\em International Conference on Medical Image Computing and
  Computer-Assisted Intervention}, pages 14--24. Springer, 2021.

\bibitem{zhang2018exfuse}
Zhenli Zhang, Xiangyu Zhang, Chao Peng, Xiangyang Xue, and Jian Sun.
\newblock Exfuse: Enhancing feature fusion for semantic segmentation.
\newblock In {\em Proceedings of the European conference on computer vision
  (ECCV)}, pages 269--284, 2018.

\bibitem{zhen2020joint}
Mingmin Zhen, Jinglu Wang, Lei Zhou, Shiwei Li, Tianwei Shen, Jiaxiang Shang,
  Tian Fang, and Long Quan.
\newblock Joint semantic segmentation and boundary detection using iterative
  pyramid contexts.
\newblock In {\em Proceedings of the IEEE/CVF Conference on Computer Vision and
  Pattern Recognition}, pages 13666--13675, 2020.

\bibitem{zheng2021rethinking}
Sixiao Zheng, Jiachen Lu, Hengshuang Zhao, Xiatian Zhu, Zekun Luo, Yabiao Wang,
  Yanwei Fu, Jianfeng Feng, Tao Xiang, Philip~HS Torr, et~al.
\newblock Rethinking semantic segmentation from a sequence-to-sequence
  perspective with transformers.
\newblock In {\em Proceedings of the IEEE/CVF conference on computer vision and
  pattern recognition}, pages 6881--6890, 2021.

\bibitem{zhou2018unet++}
Zongwei Zhou, Md~Mahfuzur Rahman~Siddiquee, Nima Tajbakhsh, and Jianming Liang.
\newblock Unet++: A nested u-net architecture for medical image segmentation.
\newblock In {\em Deep learning in medical image analysis and multimodal
  learning for clinical decision support}, pages 3--11. Springer, 2018.

\end{thebibliography}
}

\end{document}